\documentclass{article}

\usepackage{arxiv}

\usepackage[utf8]{inputenc} 
\usepackage[T1]{fontenc}    
\usepackage{hyperref}       
\usepackage{url}            
\usepackage{booktabs}       
\usepackage{amsfonts}       
\usepackage{amsmath}        
\usepackage{nicefrac}       
\usepackage{microtype}      
\usepackage{lipsum}
\usepackage{graphicx}
\graphicspath{ {./images/} }
\usepackage{cleveref} 
\crefformat{figure}{Figure~[#2#1#3]} 
\crefformat{table}{Table~[#2#1#3]} 
\crefformat{equation}{(#2#1#3)} 

\title{Out-of-Sample Hydrocarbon Production Forecasting: Time Series Machine Learning using Productivity Index-Driven Features and Inductive Conformal Prediction}

\author{
 Mohamed Hassan Abdalla Idris \\
 Sudapet Co. Ltd. \\
  University of Portsmouth \\
  Portsmouth, Hampshire PO1 2UP \\
  United Kingdom \\
  Corresponding Author: \\
  \texttt{mhdhassan@sudapet.com} \\
  \And
 Jakub Marek Cebula \\
  University of Portsmouth \\
  Portsmouth, Hampshire PO1 2UP \\
  United Kingdom \\
  \texttt{j.cebula2000@gmail.com} \\
  \And
 Jebraeel Gholinezhad \\
  University of Portsmouth \\
  Portsmouth, Hampshire PO1 2UP \\
  United Kingdom \\
  \texttt{jebraeel.gholinezhad@port.ac.uk} \\
  \And
 Shamsul Masum \\
  University of Portsmouth \\
  Portsmouth, Hampshire PO1 2UP \\
  United Kingdom \\
  \texttt{shamsul.masum@port.ac.uk} \\
  \And
 Hongjie Ma \\
  University of Portsmouth \\
  Portsmouth, Hampshire PO1 2UP \\
  United Kingdom \\
  \texttt{hongjie.ma@port.ac.uk} \\
}

\begin{document}
\maketitle
\begin{abstract}
This research introduces a new Machine Learning (ML) framework designed to enhance the robustness of out-of-sample hydrocarbon production forecasting, specifically addressing multivariate time series analysis. The proposed methodology integrates Productivity Index (PI)-driven feature selection, a concept derived from reservoir engineering, with Inductive Conformal Prediction (ICP) for rigorous uncertainty quantification. Utilizing historical data from the Volve (wells PF14, PF12) and Norne (well E1H) oil fields, this study investigates the efficacy of various predictive algorithms—namely Long Short-Term Memory (LSTM), Bidirectional LSTM (BiLSTM), Gated Recurrent Unit (GRU), and eXtreme Gradient Boosting (XGBoost)—in forecasting historical oil production rates (OPR\_H).
All the models achieved "out-of-sample" production forecasts for an upcoming future timeframe. Model performance was comprehensively evaluated using traditional error metrics (e.g., MAE) supplemented by Forecast Bias and Prediction Direction Accuracy (PDA) to assess bias and trend-capturing capabilities. The PI-based feature selection effectively reduced input dimensionality compared to conventional numerical simulation workflows. Crucially, uncertainty quantification was addressed using the ICP framework, a distribution-free approach that guarantees valid prediction intervals (e.g., 95\% coverage) without reliance on distributional assumptions, offering a distinct advantage over traditional confidence intervals, particularly for complex, non-normal data. Results demonstrated the superior performance of the LSTM model, achieving the lowest MAE on test (19.468) and genuine out-of-sample forecast data (29.638) for well PF14, with subsequent validation on Norne well E1H confirming its effectiveness. These findings highlight the significant potential of combining domain-specific knowledge with advanced ML techniques and distribution-free uncertainty quantification to improve the reliability of hydrocarbon production forecasts, thereby offering substantial implications for optimizing reservoir management and operational planning within the oil and gas industry.

\end{abstract}


\section{Introduction}
\label{sec1}

The oil and gas industry is integral to the global economy and energy demand, with accurate hydrocarbon production forecasting being crucial for strategic planning, investment decisions, and resource management. Hydrocarbons, primarily oil and gas, have been central to industrialization and economic growth for over a century, significantly influencing geopolitical relations and technological innovations. In the current era, marked by a shift towards renewable energy, the oil and gas industry faces challenges from fluctuating market demands and environmental concerns, making accurate hydrocarbon forecasting not just a technical necessity but a strategic imperative. Common units for production rates include barrels per day (bbl/day) or cubic meters per day (m3/day), and pressures are often measured in pounds per square inch (psi) or kilopascals (kPa). This context sets the stage for leveraging advancements in Machine Learning (ML) to enhance the accuracy of production predictions over typical forecasting horizons, often spanning several months to years.

Historically, hydrocarbon forecasting often involved a mix of empirical methods and expert judgment, which inherently carried uncertainties. These methods, sometimes deterministic in nature, could be constrained by linear assumptions and limited data processing capabilities. However, hydrocarbon production data, characterized by its non-linearities, cyclic and declining nature, and numerous influencing factors, often defies simplistic approaches. This study aims to address these challenges by exploring ML techniques.

In recent decades, the results of ML applications have shown potential in addressing complex, non-linear problems using relatively simple datasets across various industries. It has proved its capability to decipher inherent patterns from various datasets, extracting underlying information, efficiently \cite{Wang2023}. They present a promising solution to the hydrocarbon forecasting and associated uncertainties.

Recently, the nonlinear properties of crude oil time series have been effectively captured by models that integrate wrapper-based feature selection approaches using multi-objective optimization techniques \cite{Karasu2020}. Deep learning models, especially those that employ attention-based encoder-decoder frameworks, have shown superior forecasting performance in multivariate time series datasets \cite{Du2020}.

The datasets from Volve and Norne fields, located in the Norwegian Continental Shelf of North Sea, contain comprehensive temporal measurements of fluid production (oil, gas, and water), and wells' bottomhole pressures, serve as a perfect real-world data for conducting research on production forecasting and reservoir performance evaluation.

A significant challenge often discussed in petroleum engineering forecasting is the emphasis on predictive accuracy over genuine forecasting ability. While many studies focus on retrospective analysis using train-test splits, this work targets genuine forward-looking forecasts (i.e., out-of-sample); therefore, this study focuses explicitly on evaluating out-of-sample performance. Advancements, such as the adaptive multivariate approach, have been developed to simplify the analysis process and improve accuracy in time series forecasting \cite{Bretschneider1982}.

Uncertainty quantification remains a pivotal aspect of forecasting, regardless of prediction accuracy. This study introduces Inductive Conformal Prediction (ICP), a modern, data-driven alternative to traditional confidence intervals (CIs). CPI provides two critical advantages: (1) it guarantees valid coverage (e.g., 95\% of future observations lie within the interval) - without assuming a specific data distribution - and (2) it adapts to complex, nonlinear relationships in hydrocarbon production data. These properties address longstanding challenges in reservoir modeling, where assumptions of normality or linearity often fail. The uncertainty quantification must be dealt more thoroughly and cautiously. Floris \cite{Floris2001} has demonstrated that the quantification of uncertainty in production forecasts in reservoir and simulation models is influenced by the selection of parameterization and initial reservoir models, despite being conditioned on both static and dynamic well data, whereas the choice of reservoir simulator introduces a bias in the predicted range. In this research the importance was brought to forecast predictions as well as the underlying associated uncertainties, motivating the exploration of CPI/ICP as an alternative approach to address these limitations.

The concept of CPI for hydrocarbon production forecasting represents an emerging area. While direct applications to hydrocarbon production might be sparse in readily available literature, the principles of CPI have been explored in other dynamic time-series contexts, suggesting potential applicability here.

This paper aims to address several research gaps:
\begin{itemize}
    \item Achieving genuine out-of-sample (forward-looking) production forecasts using ML models.
    \item Investigating Productivity Index (PI) driven feature selection as a method for dimensionality reduction compared to conventional simulation approaches.
    \item Introducing and evaluating the use of Inductive Conformal Prediction (ICP) for robust uncertainty quantification in this domain, as an alternative to traditional CI.
    \item Assessing the performance of various ML models trained with these PI-driven features.
\end{itemize}

\section{Methodology}
\label{sec2}

\subsection{Productivity Index Driven Feature Selection}
\label{subsec}

Initial scrutiny involved exporting the Volve and Norne wells and the dynamic simulation parameters from the reservoir simulation model using Petrel E\&P Software \copyright \cite{SchlumbergerPetrel}, followed by the preprocessing of the dataset, and restructuring it temporally on daily basis using the KNNImputer function \cite{Troyanskaya2001} of the sklearn library in Python Programing Language.

Performance Index driven feature selection, using the related variables, proved invaluable. Ratios like the Productivity Index (PI) were utilized, to provide ML models with constraints and focus on more informative data, with the aim to provide the ML algorithms with the minimum number of variables that could potentially unearth hidden patterns, achieving significant dimensionality reduction compared to traditional simulation inputs. PI-driven features encode reservoir physics, enabling ML models to learn nonlinear patterns while maintaining interpretability.  

The PI is the pivotal metric in reservoir engineering that measures the ability of a reservoir to produce hydrocarbons across the well. In this study, the feature selection is based on the PI concept (Equation \cref{eq:pi}).

\begin{equation} \label{eq:pi}
PI = \frac{Q}{P_{\text{res}} - P_{\text{wf}}}
\end{equation}

where
\begin{itemize}
    \item $PI$ is the Productivity Index in (bbl/day)/psi or (m3/day)/kPa. Here, bbl denotes barrels.
    \item $Q$ is the production rate in (bbl/day or m3/day).
    \item $P_{res}$ is the average reservoir pressure in psi or kPa.
    \item $P_{wf}$ is the flowing bottom-hole pressure in psi or kPa.
\end{itemize}

Additionally, the PI ties closely to Darcy's Law \cite{InayatHussain1995}, serving as a metric that integrates several reservoir variables that influence the inflow performance relationship (IPR). While Darcy's Law traditionally presents a linear relationship between flow rate and pressure gradient, PI provides a composite measure that reflects the combined various nonlinear reservoir variables on production performance. In the IPR, the relationship between oil production and flowing bottom hole pressure, produces the flow characteristic curve or the flow dynamic curve \cite{Fang2014}. The inflow performance curve covers all the pressure ranges from initial reservoir pressure until full depletion i.e., zero value pressure. Being able to quantify this relation at early stages of the well life, results in an initial generalized production forecasting.

Real-world reservoir conditions, often exhibit non-linearities. A major phenomenon causing non-linearities is the reservoir depletion to below bubble point pressure (Pb). Consequently, the fluid properties, and the flow behavior undergo significant changes, altering the phase and the reservoir's fluid dynamics, leading to multiphase flow. This multiphase flow inherently complicates the PI calculation due to the changing fluid properties and relative permeabilities of oil and gas. Hence empirical PI models, like Vogel's and Fetkovich, \cite{Vogel1968}, and \cite{Fetkovich1973}, have been developed to capture these intricacies.

During the well production life, water break through at later stages, Wiggins presented a Vogel\'s like IPR models generalized for three-phase flow (oil, gas, and water) based on simulation studies \cite{Wiggins1994} (\cref{eq:wiggins_oil,eq:wiggins_water}).

\begin{equation} \label{eq:wiggins_oil}
\frac{Q_o}{Q_{o,\text{max}}} = 1 - 0.52 \left( \frac{P_{wf}}{P_{res}} \right) - 0.48 \left( \frac{P_{wf}}{P_{res}} \right)^2
\end{equation}

\begin{equation} \label{eq:wiggins_water}
\frac{Q_w}{Q_{w,\text{max}}} = 1 - 0.72 \left( \frac{P_{wf}}{P_{res}} \right) - 0.28 \left( \frac{P_{wf}}{P_{res}} \right)^2
\end{equation}

where
\begin{itemize}
    \item $Q_{o,\text{max}}$ is the maximum oil flow rate potential in (bbl/day or m3/day).
    \item $Q_{w,\text{max}}$ is the maximum water flow rate potential in (bbl/day or m3/day).
    \item $Q_o$ is the oil flow rate in (bbl/day or m3/day). 
    \item $Q_w$ is the water flow rate in (bbl/day or m3/day).
\end{itemize}

These different equations follow the same concept: calculating the ratio between the hydrocarbon production rate versus the associated differential pressure drop. They vary based on the fluid phase region (single or multi-phase) and pressure region (below or above Pb). They are quite useful and practical in determining the hydrocarbon production rate (Q) and $P_{wf}$ and enable engineers to estimate the average $P_{res}$. Accordingly, the $Q_{o,\text{max}}$ could be estimated, thus allowing estimates of the production rates for different $P_{wf}$ at the same average reservoir pressure $P_{res}$.

The Wiggins IPR introduces the gas, oil, and water production ratios, aiming to capture the complexities and non-linearities inherent in hydrocarbon production. However, traditional models like the Wiggins IPR often fall short in accurately predicting production due to their limited ability to adapt to the intricate and dynamic nature of reservoir behavior and the whole production process. These models typically rely on predefined equations and assumptions that may not fully capture the variability and interactions in the data. ML, with its inherent capability to model non-linear relationships and decipher hidden patterns in large datasets, offers a promising avenue for enhancing the accuracy of these forecasting endeavors \cite{Wu2020}. ML models can learn from historical data, adapt to changing conditions, and uncover complex patterns that traditional models might miss, leading to more robust and reliable predictions.

The features selection criteria are PI driven. The Volve field average reservoir pressure ($P_{res}$) was supported via the water and gas injection scheme, and as well it was maintained above the Bubble Point Pressure (Pb) according to the field Plan for Development and Production (PDO) report \cite{Equinor2018}, therefore the Bottom-hole Flowing Pressure ($P_{wf}$) is assumed to solely drive the pressure differential ($\Delta$P) changes throughout the well production lifetime. Accordingly, the following features were used in the machine learning models:
\begin{itemize}
    \item OPR\_H: Historical Daily Oil Production Rate, used as target feature.
    \item WPR\_H: Historical Daily Water Production Rate.
    \item BHP\_H: Historical Daily Bottom-hole Flowing Pressure ($P_{wf}$). 
\end{itemize}

The reservoir simulation results of the field were used for forward-looking (out-of-sample) forecasting of the daily production for future out-of-sample 10 months horizon for PF14 and PF12 of Volve field and 12 months for E1H of Norne Field. As well the forecasting results were compared to the simulation results and against the actual historical oil production. The simulated features used for this forecasting period were:
\begin{itemize}
    \item OPR: Simulated Daily Oil Production Rate.
    \item GPR: Simulated Daily Gas Production Rate.
    \item WPR: Simulated Daily Water Production Rate.
    \item BHP: Simulated Daily Bottom-hole Flowing Pressure ($P_{wf}$).
\end{itemize}

The reservoir conditions per the field PDO report \cite{Equinor2018}, stated that reservoir pressure is 340 bar (equivalent to 4931.3 psi).

\subsection{Data Splitting}
\label{subsec2}

Excluding the targeted out-of-sample forecasting horizon which was not fed into the model, the dataset was split into training and testing subsets. The training dataset, constituting 80\% of the total data, was used for training the models, the remaining 20\% was reserved for validation and testing purposes. This approach ensures that the models are generalizable and not just tailored for one specific dataset. Upon successful modelling of well PF-14, the models are then validated on new well PF-12. The updated model is then  validated on well E1H sourced from the Norne Oil Field.

\subsection{Challenges in Time Series Analysis}
\label{subsec3}

Despite its potency, time series analysis isn't devoid of challenges. The inherent autocorrelation in time series data can lead to misleading statistical inferences. Moreover, structural breaks due to internal, and external shocks like well intervention, power trips in wells' pumps, policy changes, sudden global events, or technological advancements can disrupt established patterns and shifts in data behavior, making predictions challenging.

\begin{figure}[htb!]
\centering
\includegraphics[width=\linewidth]{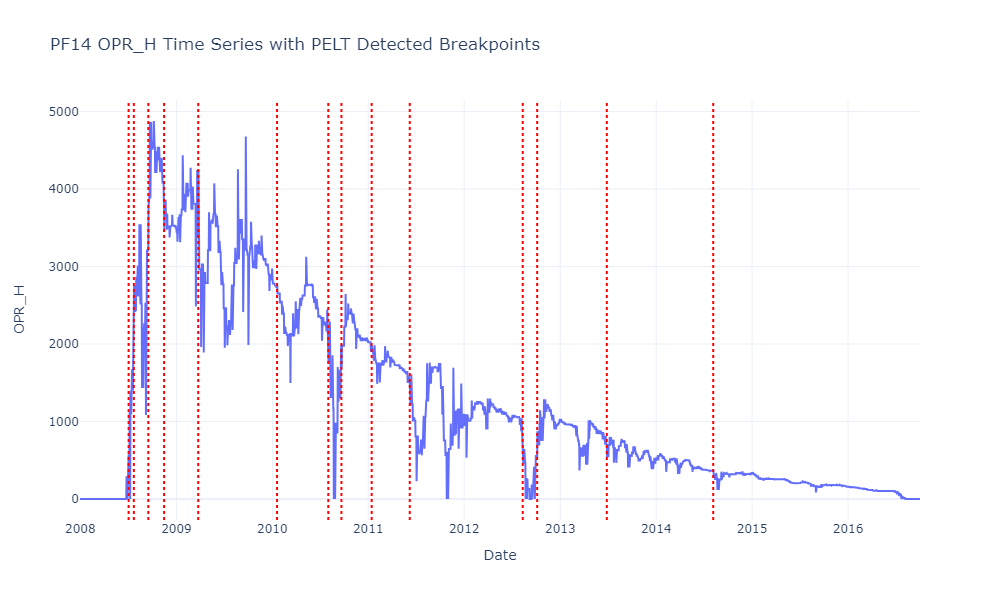}
\caption{Pelt Detected Breakpoints, PF14 OPR\_H}
\label{fig:figure1}
\end{figure}

\begin{figure}[htb!]
\centering
\includegraphics[width=\linewidth]{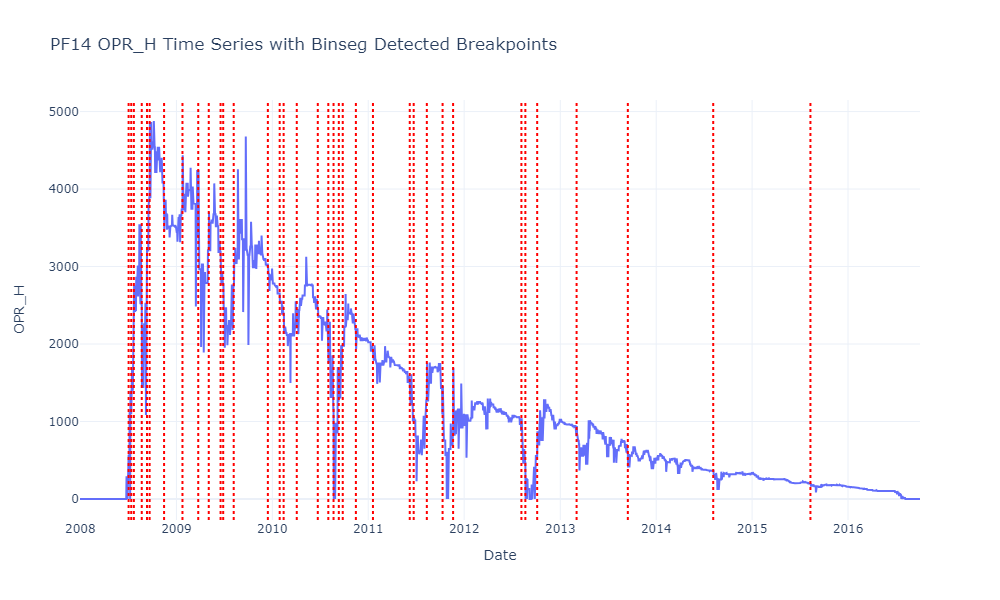}
\caption{Binseg Detected Breakpoints, PF14 OPR\_H}
\label{fig:figure2}
\end{figure}

The "Change Point Detection" test, which identifies times when the statistical properties of a time series change, was carried out by applying the "Pelt Algorithm" \cite{Killick2012}, an efficient method for exact segmentation based on minimizing a cost function. This was chosen for its efficiency in finding an unknown number of change points. Equally, the "Binseg Algorithm" \cite{Bai1997}, a binary segmentation approach for detecting multiple changepoints, with its recursive approach for finding significant change points was applied. The tests resulted in Figures \cref{fig:figure1,fig:figure2}, leading to observed discrete structural breaks. These points indicate interruption events that took place during production i.e., well intervention activities or power trip in the wells. This shows how the production data in its temporal nature behaves differently throughout the wells' lifetime.

\subsection{Models for Production Forecasting}\label{subsec4}

\begin{figure}[htb!]
\centering
\includegraphics[width=\linewidth]{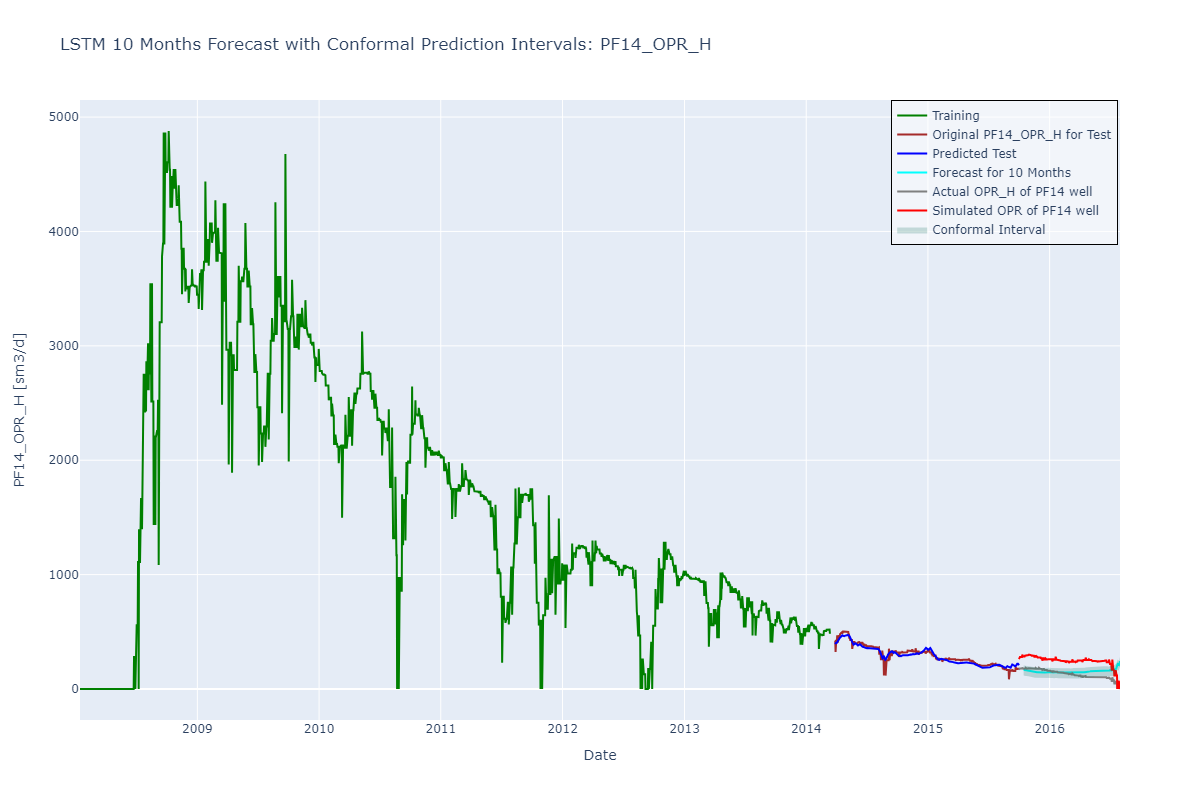}
\caption{LSTM 10 months out-of-sample forecast for PF14 well\_H}
\label{fig:figure3}
\end{figure}

In hydrocarbon production forecasting, where the data is multivariate in nature and with inherent complexities, ML proves instrumental. Unlike traditional time series forecast methods, ML offers the capability to identify hidden patterns in these high-dimensional datasets, search for complex and non-linear relationships, and avoid the need for preliminary definitions of mathematical relations among the model's input data.
The four primary models covered were: Long Short-Term Memory (LSTM), Bidirectional LSTM (BiLSTM), Gated Recurrent Unit (GRU), and eXtreme Gradient Boosting (XGBoost). The rationale is that these ML models are specifically designed to capture complex correlations within data without the need for explicit programming, with the capacity for adaptation enabling ML to effectively capture non-linear patterns, interactions between variables, and complex relationships that may defy conventional models and remain unaccounted for.
The PI driven feature selection approach allowed optimization for the best, yet low dimension set of variables (only 3 input features plus the target) that proved to have a high impact on forecasting production. Three wells were used in this study: PF14 and PF12 of Volve field and E1H of Norne field for validation. The reason for choosing these wells is their long production history covering the whole field lifetime. The Volve field time spans from 15-01-2008 to 30-07-2016, whereas Norne field is 06-11-1997 to 01-12-2006. The ML algorithms were trained for forecasting, with the train-test split of 80:20 for all models. The data frame (df) for model train-test split was set from 15-01-2008 to 01-10-2015 for the Volve wells and 06-11-1997 to 01-12-2005 for E1H of Norne.
The period of 01-10-2015 to 30-07-2016 and 01-12-2005 to 01-12-2006 was set as out-of-sample for blind forecasting for Volve and Norne wells respectively, using the simulated data ['WPR', 'GPR', 'BHP'] as inputs, instead of the historic ['WPR\_H', 'GPR\_H', 'BHP\_H']. This allowed testing blind forecast performance for the given forecast interval (i.e., forecast horizon), and also allowed comparison of these results against the simulator forecasting results as well as the actual production historical readings. Following the forecast ML modeling on PF14 well of Volve field, the four models were cross validated on PF12 well of the same field. After comparison and assessment of all models for both wells, the best performing model (LSTM) was further validated on the E1H well of Norne field.
The Norne was discovered in 1991 and its development began later in August 1996. The first oil production was achieved a year later in November 1997. Operated by Statoil, lately it merged as Equinor in May 2018 \cite{Rwechungura2010}. Norne oil is produced with water injection recovery mechanism, following the same pressure support concept as of the Volve Field and was and still is operated by same operator, Equinor.
The data set of Norne retrieved from the 'The Open Porous Media (OPM)' website. The dataset is licensed under the Open Database License (ODbL) \cite{OPMProject2023}. The Reservoir Dynamic Model contained production data from 1997 to 2006.
Furthermore, as both production datasets were exported from the Schlumberger Petrel 2016 Software, the preparation, and the way they were laid out was identical, but Norne had no historic bottom hole flowing pressure recorded, hence the BHP\_H was substituted by the simulated BHP which suffices the purpose of this study.
Hyperparameters' tuning, using grid search and random search techniques was carried on all models for best hyperparameters to optimize model performance. The period for the forecast in PF14 and PF12 of Volve was set from October 1st, 2015, until July 30th, 2016, which is equal to 10 months and 11 days (approximately 0.83 years), and for E1H of Norne was from December 1st, 2005, until December 1st, 2006 (1 year). Training robust models demands meticulous attention to data preparation, model architecture, and validation strategy. Cross-validation, particularly time-series cross-validation, becomes crucial to avoid lookahead biases and to provide an honest evaluation of the model's potential performance on unseen data, where CPI is introduced.
For each ML model—whether LSTM, BiLSTM, GRU, or XGBoost—hyperparameters were tuned to optimize performance, ensuring that the model neither overfits to the training data nor underfits, missing out on crucial patterns. 
The model train-test split and out-of-sample forecast for PF14 well with its CPI versus both actual and simulation production results is shown in Figure \cref{fig:figure3}, representing LSTM.

Forecast error and model performance were gauged using different metrics to assess effectiveness. The Mean Absolute Error (MAE) offers an understanding of the average magnitude of errors, irrespective of their direction. The Root Mean Squared Error (RMSE), by penalizing larger errors more heavily, provides insights into the volatility of the forecast predictions. The Symmetric Mean Absolute Percentage Error (sMAPE) offers a scale-independent metric, making it invaluable for comparing performance across different wells or regions. Additionally, the Forecast Bias provides a measure of systematic over or under-predictions, while the Prediction Direction Accuracy (PDA) reveals the model's capability in capturing trends. Together, these metrics provide a comprehensive view of the model's forecasting abilities. A reasonable level of accuracy in this context would typically involve minimizing these error metrics (MAE, RMSE, sMAPE) while maintaining low bias and high PDA, though specific acceptable thresholds can vary depending on the application.

The Forecast Bias metric (Equation \cref{eq:forecast_bias}) gives an idea about the direction of the error made by the model. A positive value indicates that forecasts are on average too high, while a negative value signifies that they are too low. Whereas the PDA metric calculates the percentage of times the model correctly predicts the direction of change in the variable; the calculation for this metric isn't a standard formula like the others, as it's more of a conditional counting operation.

\begin{equation} \label{eq:forecast_bias}
    \text{Forecast Bias} = \frac{1}{n} \sum_{i=1}^{n} (y_i - \hat{y}_i)
\end{equation}

where
\begin{itemize}
    \item $y_i$ is the actual value.
    \item $\hat{y}_i$ is the predicted value.
    \item $n$ is the number of observations.
\end{itemize}

To quantify uncertainty, we employed Inductive Conformal Prediction (ICP), a scalable variant of conformal prediction. ICP constructs prediction intervals with finite-sample validity, ensuring (e.g., 95\%) coverage regardless of the data distribution. This approach contrasts with traditional confidence intervals, which rely on parametric assumptions (e.g., normality) and fail to account for nonlinearities or heteroscedasticity in hydrocarbon production data.  \Cref{eq:cpi} represents a symmetric interval around the point prediction $\hat{y}_i$, where $\epsilon$ is the half-width of the interval.
The significance of this technique cannot be overstated, given the unpredictable dynamics of hydrocarbon production; it's crucial to have forecasts that not only pinpoint accuracy but also offer a defined level of confidence \cite{Shafer2007}. This is where inductive conformal prediction shines, as it delivers prediction intervals backed by a coverage probability assurance. These intervals are crafted to encompass the genuine value of the intended variable at a set probability, independent of the data distribution. Conformal prediction offers a paradigm shift for uncertainty quantification in hydrocarbon forecasting. 

\begin{equation} \label{eq:cpi}
    \text{Inductive Conformal Prediction (ICP)} = [\hat{y}_i - \epsilon, \hat{y}_i + \epsilon]
\end{equation}

Equation \cref{eq:cpi} represents a symmetric interval around the point prediction $\hat{y}_i$, where $\epsilon$ is the half-width of the interval.

\begin{itemize}
    \item $\hat{y}_i$ represents the point estimate or prediction for the $i$-th observation.
    \item $\epsilon$ is the margin of error around the point prediction; its size determines the width of the prediction interval.
    \item $[\hat{y}_i - \epsilon, \hat{y}_i + \epsilon]$ is the prediction interval. For any given observation, there's a specified probability (e.g., 95\%) that the true value of the target variable will fall within this interval.
\end{itemize}

A Confidence Interval (CI) gives an estimated range of values which is likely to include an unknown population parameter. It is constructed using sampled data and provides a range within which we expect the true parameter to lie, with a certain level of confidence. The width of the interval gives us an idea of how uncertain we are about the unknown parameter. A smaller interval might indicate more confidence that the parameter lies within the interval, whereas a larger interval might show less confidence.
Conformal Prediction (CP), on the other hand, provides a framework for creating prediction intervals that attain a desired level of coverage under any distributional assumption. The method is distribution-free in the sense that it makes no probabilistic assumptions about the data-generating process. Instead, it uses the empirical distribution of the data to provide valid prediction intervals. The key advantage of CP is its robustness to model misspecification. Conformal prediction is a technique used in machine learning to predict a range of potential values (an interval) instead of a single value. It provides a measure of the uncertainty associated with predictions.
Advancements in conformal predictors have led to the development of conformal predictive framework. Unlike traditional ones that offer set predictions in regression challenges, this framework provides probability distributions for test observation labels. In this study, the Inductive Conformal Prediction (ICP) \cite{Vovk2019} is implemented. ICP involves splitting the data into two sets: a proper training set and a calibration set. The models in this research are calibrated on the test set (20\% of the historical data), and nonconformity scores are computed using the residuals. This approach is more computationally efficient than full conformal prediction.
After all residuals on the calibration set are calculated, the code computes the quantile corresponding to the desired coverage level (e.g., 95\% coverage implies using the $1 - \alpha = 0.95$ quantile of residuals, where $\alpha=0.05$). This quantile determines the margin $\epsilon$. In this context, these bounds are used to create a 95\% prediction interval, meaning that under the exchangeability assumption between calibration and test data, 95\% of the true future values are expected to lie within this interval. This is emphasized by the fact that conformal prediction uses the concept of nonconformity measures. A nonconformity measure is a function that assigns a numerical value to each example, indicating how 'strange' or non-conforming this example is, given the model trained on the proper training set. The higher the nonconformity score, the less typical the example is. Using the distribution of nonconformity scores on the calibration set, conformal prediction determines the threshold $\epsilon$ needed to achieve the target coverage probability for new test examples. The key advantages of ICP include: 
\begin{itemize}
    \item No distributional assumptions: Unlike CIs, ICP works for non-normal, heteroscedastic, or multimodal data.  
    \item Adaptability: Automatically adjusts to nonlinearities and temporal shifts (e.g., reservoir depletion).  
    \item Interpretability: Provides clear, actionable intervals for decision-making (e.g., risk assessment in drilling operations).  
\end{itemize}

\begin{figure}[htb!]
\centering
\includegraphics[width=\linewidth]{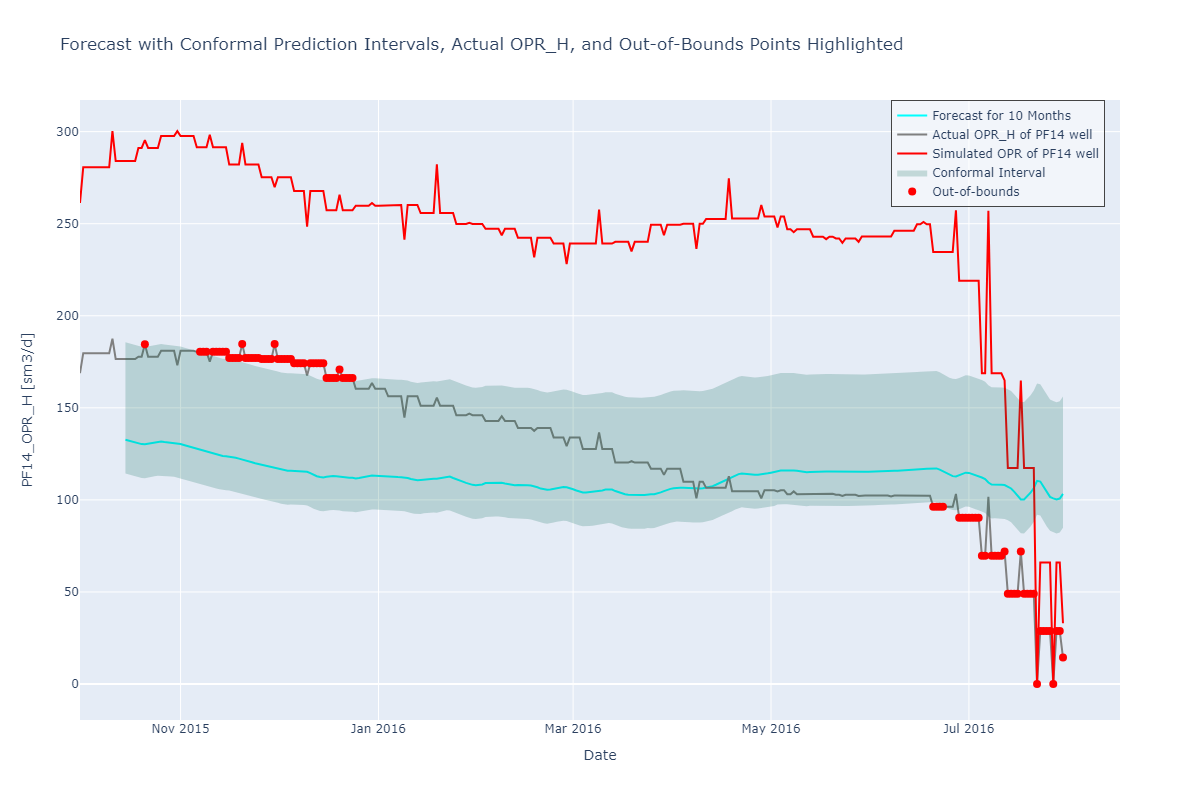}
\caption{LSTM Forecast with CP vs. Actual and Out-of-Bounds Points Well PF14 OPR\_H}
\label{fig:figure4}
\end{figure}

In the context of this study, CPI was used to provide prediction intervals for the LSTM, BiLSTM, GRU, and XGBoost models when forecasting the oil production rate. The nonconformity measure used was based on the residuals of the model on the 20\% test predictions dataset (calibration set). Using this measure, prediction intervals were generated for the test dataset, and the proportion of test examples that fell outside the prediction intervals was calculated. This proportion provides a measure of the calibration of the prediction intervals: for 95\% prediction intervals, thus about 5\% of test examples expect to fall outside the intervals. The CPI framework demonstrated strong calibration across all models. For example, the LSTM model achieved 70.34\% coverage** for PF14 (\cref{fig:figure4}), aligning closely with the target 95\% confidence level. Notably, CPI outperformed traditional confidence intervals in scenarios with non-Gaussian residuals or structural breaks (e.g., abrupt changes in production due to well interventions). This robustness underscores CPI's suitability for real-world reservoir management, where data often violates parametric assumptions. 

Similarly, Figures \cref{fig:figure6,fig:figure8} show the same for well PF12, versus actual and simulation results.

\begin{figure}[htb!]
\centering
\includegraphics[width=\linewidth]{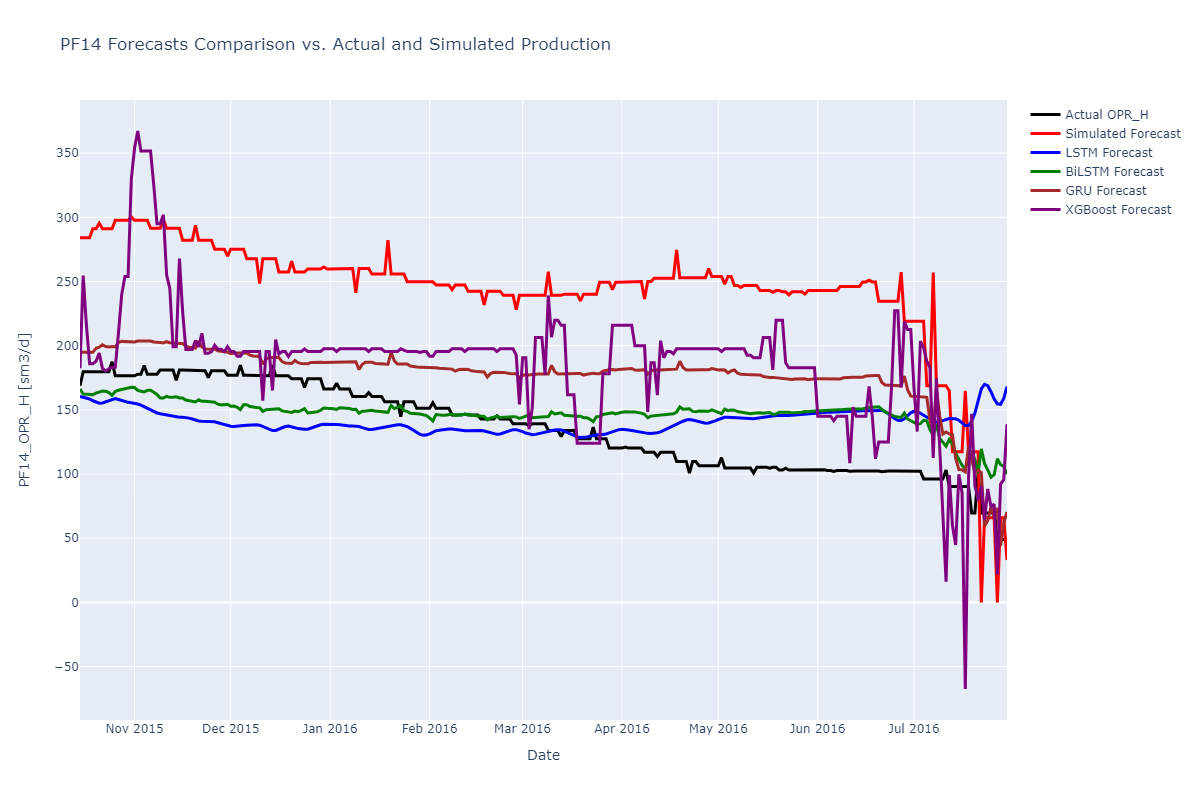}
\caption{PF14 Forecasts vs. Actual and Simulation Production}
\label{fig:figure5}
\end{figure}

\begin{figure}[htb!]
\centering
\includegraphics[width=\linewidth]{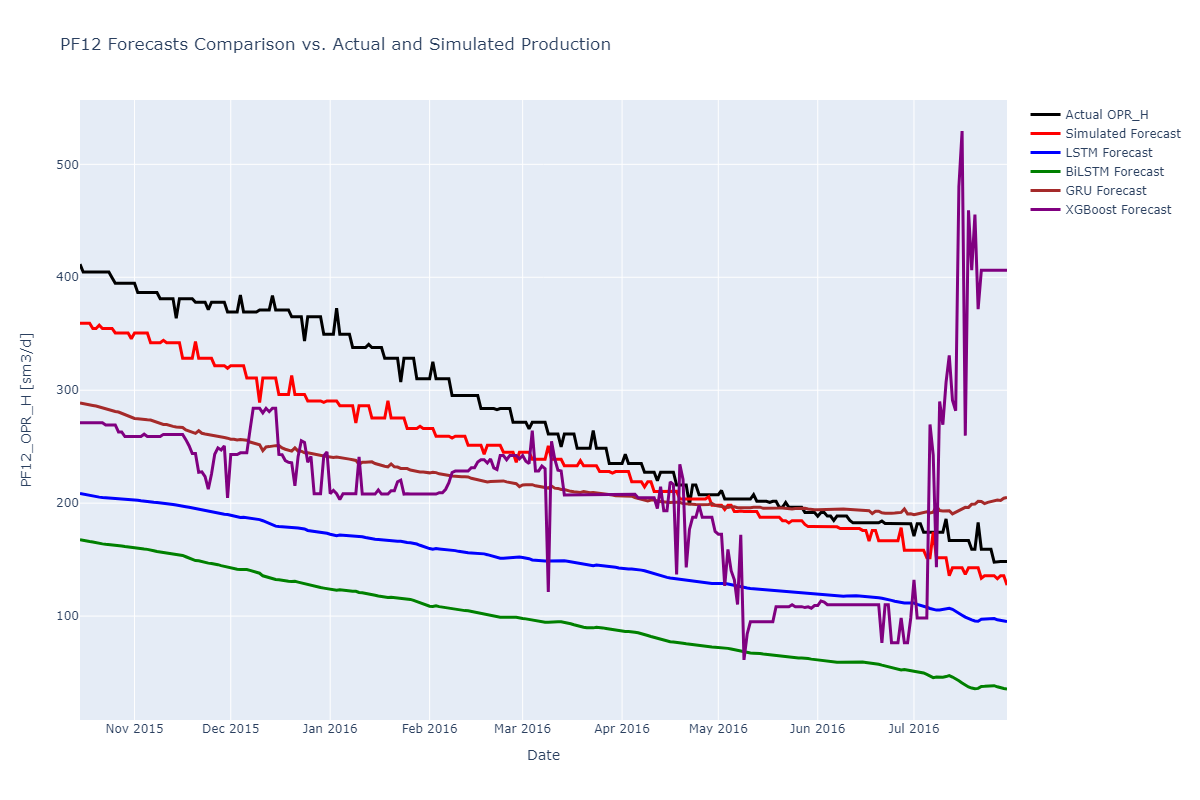}
\caption{PF12 Forecasts vs. Actual and Simulation Production}
\label{fig:figure6}
\end{figure}

\begin{figure}[htb!]
\centering
\includegraphics[width=\linewidth]{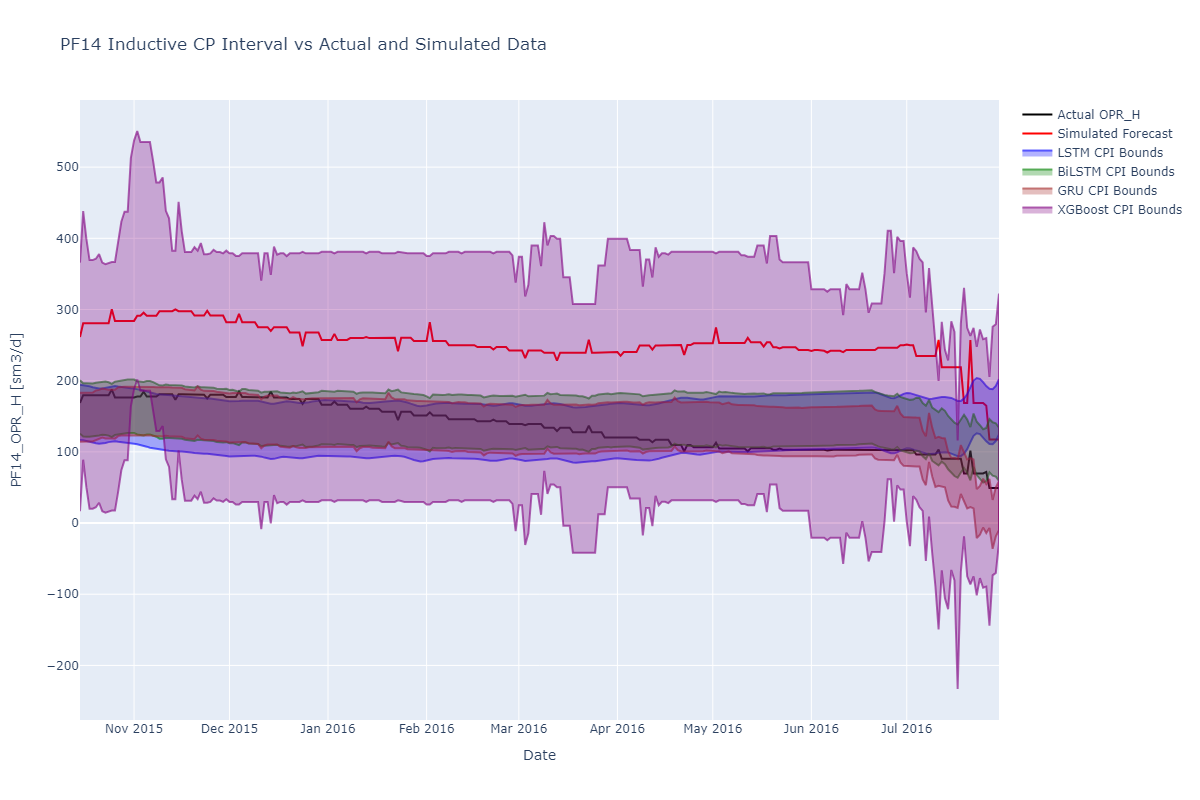}
\caption{PF14 ICP Intervals vs. Actual Production}
\label{fig:figure7}
\end{figure}

\begin{figure}[htb!]
\centering
\includegraphics[width=\linewidth]{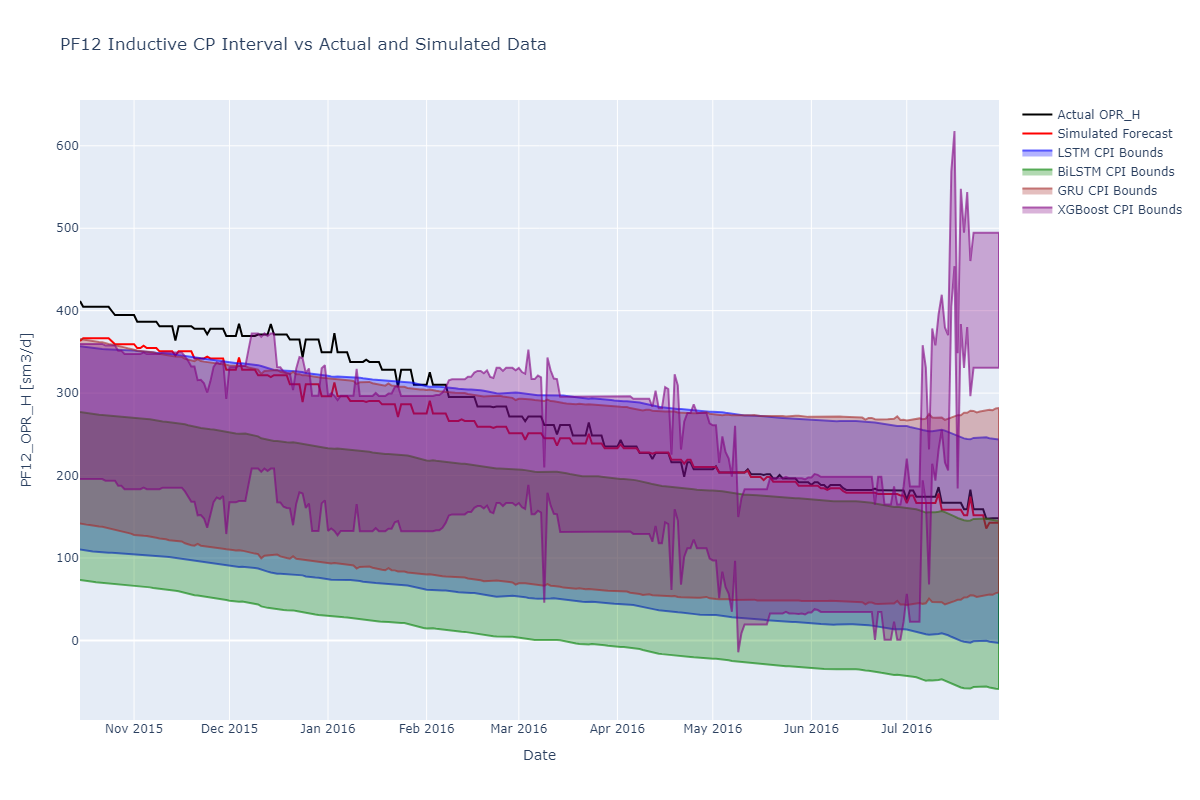}
\caption{PF12 ICP Intervals vs. Actual Production}
\label{fig:figure8}
\end{figure}

\section{Results and Discussion}\label{sec3}
\subsection{Comparative Analysis of ML Techniques – Metrics' Evaluation}\label{subsec4}

\begin{figure}[htb!]
    \centering
    \begin{minipage}{0.45\textwidth}
        \centering
        \includegraphics[width=\textwidth]{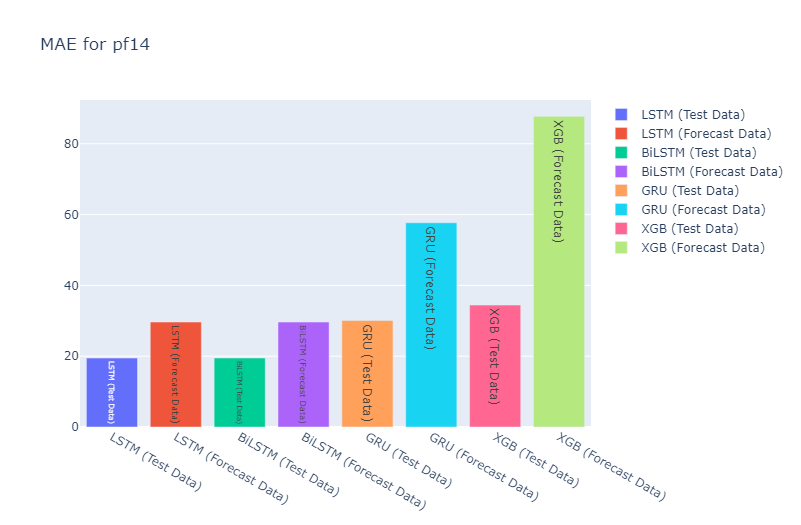}
        \caption{Performance and Error Metrics PF14}
        \label{fig:figure9}
    \end{minipage}
    \hfill
    \begin{minipage}{0.45\textwidth}
        \centering
        \includegraphics[width=\textwidth]{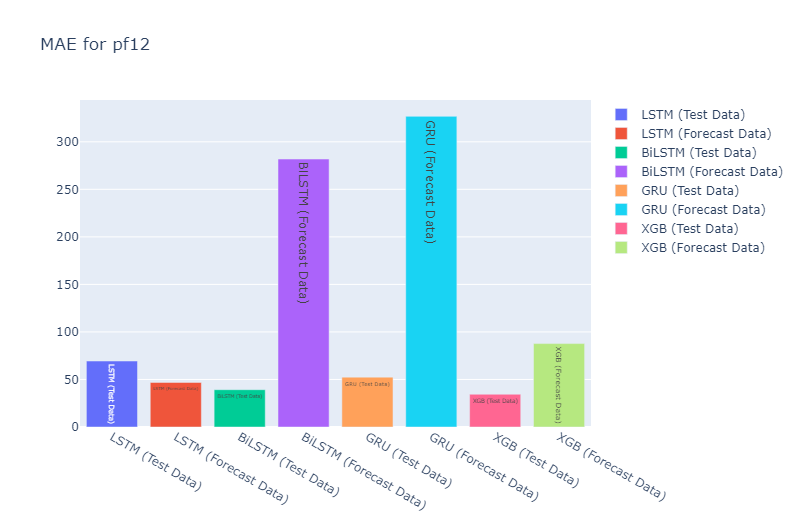}
        \caption{Performance and Error Metrics PF12}
        \label{fig:figure10}
    \end{minipage}
\end{figure}

\begin{figure}[htb!]
    \centering
    \begin{minipage}{0.45\textwidth}
        \centering
        \includegraphics[width=\textwidth]{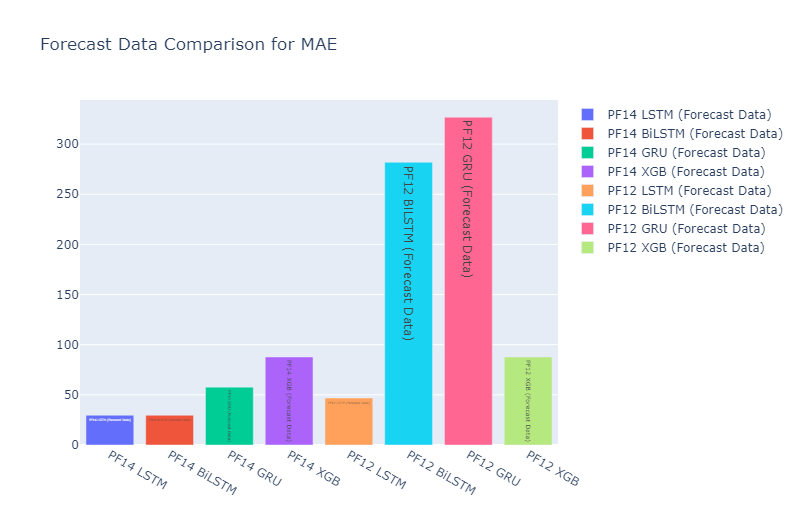}
        \caption{MAE Error Metrics PF14 and PF12 Forecasts}
        \label{fig:figure11}
    \end{minipage}
    \hfill
    \begin{minipage}{0.45\textwidth}
        \centering
        \includegraphics[width=\textwidth]{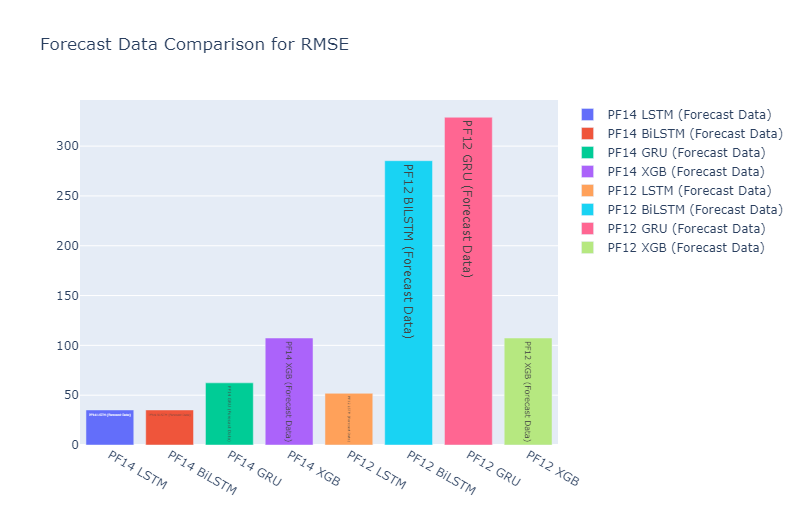}
        \caption{RMSE Error Metrics PF14 and PF12 Forecasts}
        \label{fig:figure12}
    \end{minipage}
\end{figure}

\begin{figure}[htb!]
    \centering
    \begin{minipage}{0.45\textwidth}
        \centering
        \includegraphics[width=\textwidth]{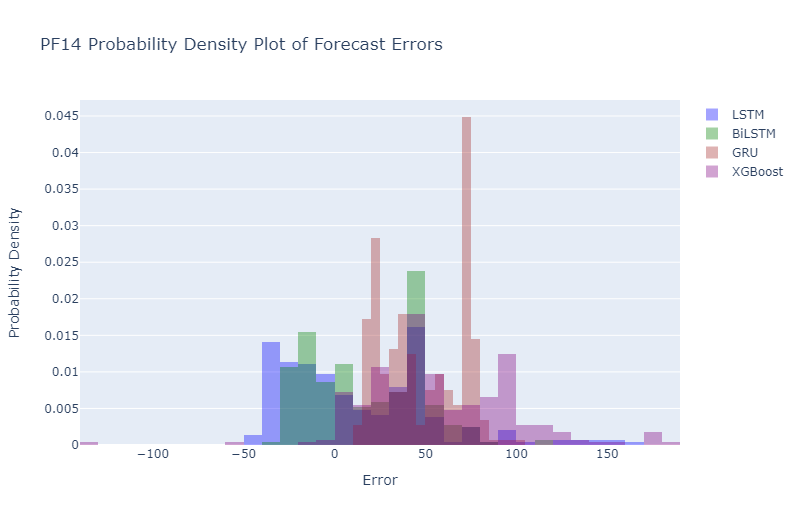}
        \caption{Forecast Errors Probability Density Plot PF14}
        \label{fig:figure13}
    \end{minipage}
    \hfill
    \begin{minipage}{0.45\textwidth}
        \centering
        \includegraphics[width=\textwidth]{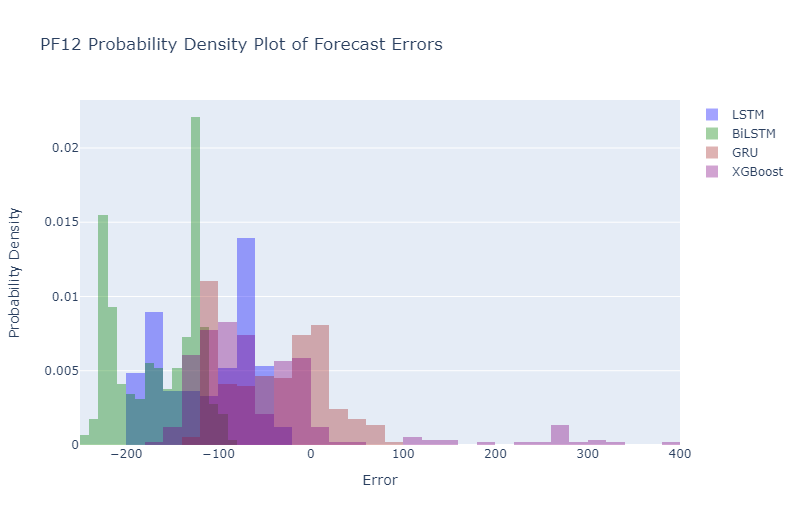}
        \caption{Forecast Errors Probability Density Plot PF12}
        \label{fig:figure14}
    \end{minipage}
\end{figure}

\begin{figure}[htb!]
    \centering
    \begin{minipage}{0.45\textwidth}
        \centering
        \includegraphics[width=\textwidth]{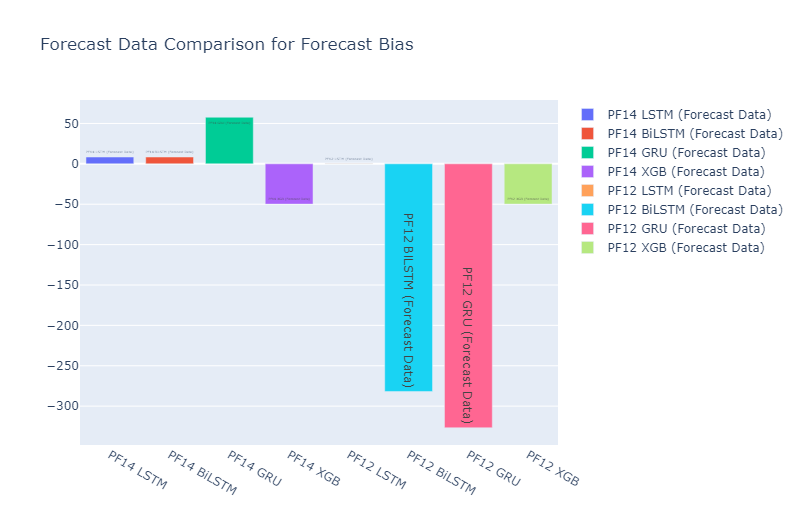}
        \caption{ML Forecast Bias PF14 vs. PF12}
        \label{fig:figure15}
    \end{minipage}
    \hfill
    \begin{minipage}{0.45\textwidth}
        \centering
        \includegraphics[width=\textwidth]{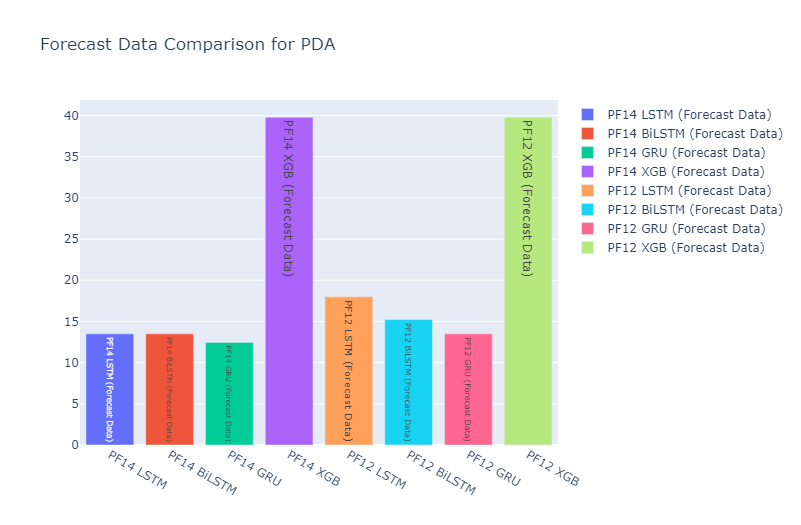}
        \caption{ML PDA PF14 vs. PF12}
        \label{fig:figure16}
    \end{minipage}
\end{figure}

\begin{figure}[htb!]
    \centering
    \begin{minipage}{0.45\textwidth}
        \centering
        \includegraphics[width=\textwidth]{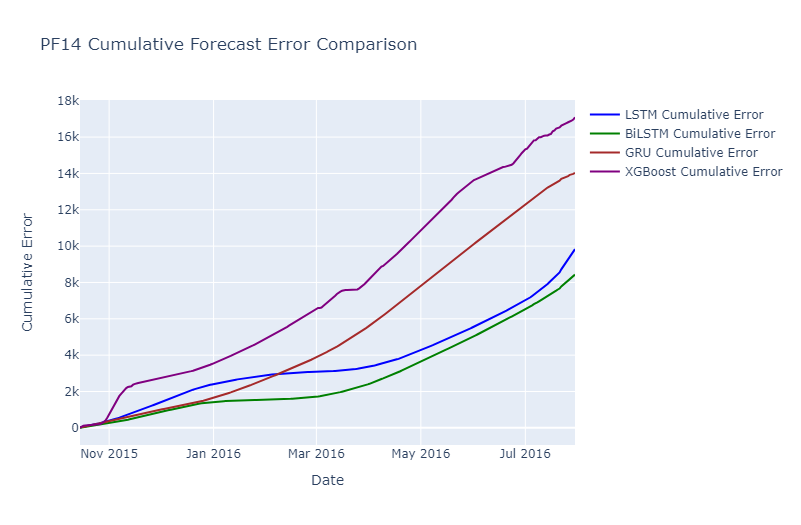}
        \caption{Cumulative Forecast Error PF14 Models}
        \label{fig:figure17}
    \end{minipage}
    \hfill
    \begin{minipage}{0.45\textwidth}
        \centering
        \includegraphics[width=\textwidth]{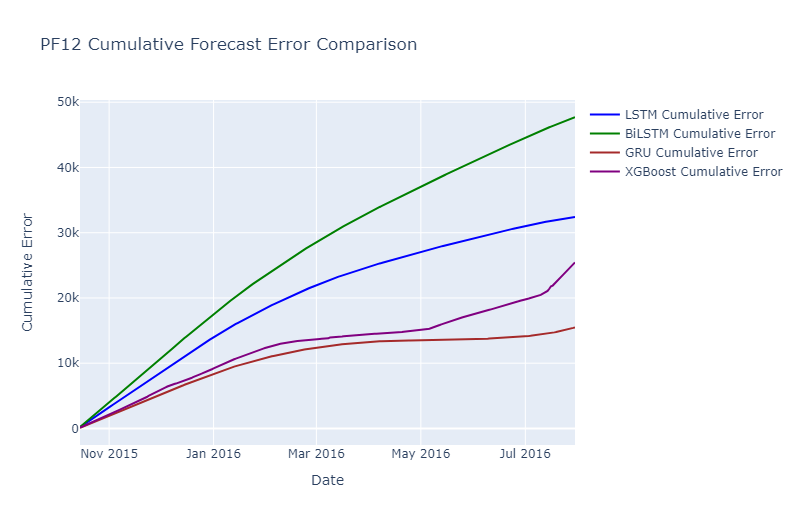}
        \caption{Cumulative Forecast Error PF12 Models}
        \label{fig:figure18}
    \end{minipage}
\end{figure}

The LSTM, BiLSTM, GRUs, and XGBoost  models were assessed based on their capability to predict the historical oil production rate (OPR\_H) using two datasets, namely wells PF14 and PF12. The evaluation metrics chosen to gauge the performance of these models consisted of three error metrics and two performance metrics. After models being fit, an "out-of-sample" blind forecasting was carried for the future horizon of 10 months, in this exercise new dataset was used i.e., the simulated features instead of the historical ones, accordingly the comparison between the ML models\' forecast and the industry simulated forecast are logical to be compared versus each other in terms of error and performance metrics, and an assessment results would be driven in order to prove or not the reliability and efficiency of ML models.(\Cref{fig:figure9,fig:figure10,fig:figure11,fig:figure12,fig:figure13,fig:figure14,fig:figure15,fig:figure16,fig:figure17,fig:figure18}).

The analysis of Mean Absolute Error (MAE) for well PF14 reveals that the LSTM model demonstrates the lowest MAE on test data (19.468), indicating it effectively captures the historical oil production rate (OPR\_H). For forecast data, LSTM still maintains a relatively low MAE (29.638) compared to GRU (57.688) and significantly better than XGB (87.742). When considering the Root Mean Squared Error (RMSE), LSTM shows the lowest values for both test (24.195) and forecast data (35.192), suggesting its robustness in handling time series data. BiLSTM has a higher RMSE (24.533 test, 43.574 forecast), highlighting potential difficulties in generalizing to unseen data compared to LSTM. In terms of sMAPE (Symmetric Mean Absolute Percentage Error), LSTM demonstrates superior performance with the lowest sMAPE for test data (6.943\\

Examining Forecast Bias, BiLSTM exhibits minimal bias in test data (-6.940), while LSTM shows a consistent negative bias (-12.295). In forecast data, BiLSTM and GRU display positive biases (12.644 and 57.688, respectively), whereas LSTM (8.438) and XGB (-49.860) maintain significant negative biases. In addition, the PDA results show that GRU and BiLSTM exhibit the best performance with low PDA values, indicating high precision in both test and forecast data. Specifically, GRU (15.942\% test, 12.457\% forecast) and BiLSTM (18.841\% test, 14.533\% forecast) outperform LSTM (13.225\% test, 13.495\% forecast) and XGB (34.601\% test, 39.792\% forecast).

LSTM and BiLSTM, both being variants of recurrent neural networks, demonstrated significant potential in handling this time series data, as expected. The XGBoost, a gradient boosting algorithm, showed a competitive performance as well given its non-sequential nature.

For well PF12, the analysis of Mean Absolute Error (MAE) indicates that BiLSTM (39.244 test, 281.845 forecast) and XGB (34.437 test, 87.742 forecast) outperform other models. LSTM shows improvement in forecast data (46.855) compared to its test data (69.393), indicating potential generalization issues or improvement in long-term forecasting. Considering Root Mean Squared Error (RMSE), XGB and BiLSTM show lower RMSE values (47.291 and 56.222 test; 107.393 and 285.368 forecast), indicating better predictive performance compared to LSTM (88.383 test, 51.877 forecast) and GRU (70.993 test, 328.970 forecast). In terms of sMAPE, BiLSTM and XGB demonstrate better performance with lower sMAPE values (20.041\% and 24.312\% test; 192.497\% and 37.547\% forecast). The high sMAPE for LSTM (29.545\% test, 18.603\% forecast) and GRU (25.835\% test, 198.744\% forecast) suggests overfitting issues.

Analyzing Forecast Bias, LSTM shows a large positive bias in test data (29.729) and a significant negative bias in forecast data (0.410), indicating inconsistent performance. BiLSTM (-7.945 test, -281.845 forecast) and XGB (-12.548 test, -49.860 forecast) show more stable bias values. In addition, the PDA results reveal that BiLSTM and GRU perform better in terms of PDA for PF12. BiLSTM (16.123\% test, 15.225\% forecast) and GRU (13.768\% test, 13.495\% forecast) outperform LSTM (18.659\% test, 17.993\% forecast) and XGB (34.601\% test, 39.792\% forecast).

The LSTM model stands out for its consistent performance across both wells and scenarios. It suggests the model's capability to effectively capture the time dependencies in the datasets, even with the challenges presented by incomplete data recording in PF12. The significant increase in error rates for BiLSTM and GRU in the forecast data, especially for PF12, might be indicating potential overfitting to the test data or challenges in generalizing to unseen data. XGB, despite being a non-sequential model, shows commendable performance, particularly in handling the PF12 dataset, indicating its effectiveness in scenarios where data dependencies are not strongly temporal. (\Cref{tab:table1,tab:table2}).
It's noteworthy that while LSTM performed admirably in tests for both wells, its performance on the forecast data for PF14 was particularly commendable. This suggests that LSTM's memory cells can effectively capture the time dependencies of the PF14 dataset. Considering that the BHP\_H of PF12 was not completely recorded throughout the well's lifetime, this further approves the effectiveness and robustness of LSTM and other recurrent neural networks.

\begin{table}[h]
\centering
\caption{Error and Performance Metrics PF14}
\label{tab:table1}
\resizebox{\textwidth}{!}{%
\begin{tabular}{ccccccccc}
\hline
\textbf{PF14 Metrics} &
  \textbf{LSTM (Test Data)} &
  \textbf{LSTM (Forecast Data)} &
  \textbf{BiLSTM (Test Data)} &
  \textbf{BiLSTM (Forecast Data)} &
  \textbf{GRU (Test Data)} &
  \textbf{GRU (Forecast Data)} &
  \textbf{XGB (Test Data)} &
  \textbf{XGB (Forecast Data)} \\ \hline
\textbf{MAE}           & 19.468  & 29.638 & 19.815 & 33.906 & 30.084 & 57.688 & 34.437  & 87.742  \\ \hline
\textbf{RMSE}          & 24.195  & 35.192 & 24.533 & 43.574 & 35.842 & 62.378 & 47.291  & 107.393 \\ \hline
\textbf{sMAPE}         & 6.943   & 25.489 & 7.527  & 27.872 & 11.449 & 41.302 & 24.312  & 37.547  \\ \hline
\textbf{Forecast Bias} & -12.295 & 8.438  & -6.940 & 12.644 & 24.498 & 57.688 & -12.548 & -49.860 \\ \hline
\textbf{PDA}           & 13.225  & 13.495 & 18.841 & 14.533 & 15.942 & 12.457 & 34.601  & 39.792  \\ \hline
\end{tabular}%
}
\end{table}

\begin{table}[h]
\centering
\caption{Error and Performance Metrics PF12}
\label{tab:table2}
\resizebox{\textwidth}{!}{%
\begin{tabular}{ccccccccc}
\hline
\textbf{PF12 Metrics} &
  \textbf{LSTM (Test Data)} &
  \textbf{LSTM (Forecast Data)} &
  \textbf{BiLSTM (Test Data)} &
  \textbf{BiLSTM (Forecast Data)} &
  \textbf{GRU (Test Data)} &
  \textbf{GRU (Forecast Data)} &
  \textbf{XGB (Test Data)} &
  \textbf{XGB (Forecast Data)} \\ \hline
\textbf{MAE}           & 69.393 & 46.855 & 39.244 & 281.845  & 52.454 & 326.755  & 34.437  & 87.742  \\ \hline
\textbf{RMSE}          & 88.383 & 51.877 & 56.222 & 285.368  & 70.993 & 328.970  & 47.291  & 107.393 \\ \hline
\textbf{sMAPE}         & 29.545 & 18.603 & 20.041 & 192.497  & 25.835 & 198.744  & 24.312  & 37.547  \\ \hline
\textbf{Forecast Bias} & 29.729 & 0.410  & -7.945 & -281.845 & 15.088 & -326.755 & -12.548 & -49.860 \\ \hline
\textbf{PDA}           & 18.659 & 17.993 & 16.123 & 15.225   & 13.768 & 13.495   & 34.601  & 39.792  \\ \hline
\end{tabular}%
}
\end{table}

\subsection{Norne Field – Model Cross Validation and Deployment Case}\label{subsec5}

\begin{figure}[htb!]
\centering
\includegraphics[width=\linewidth]{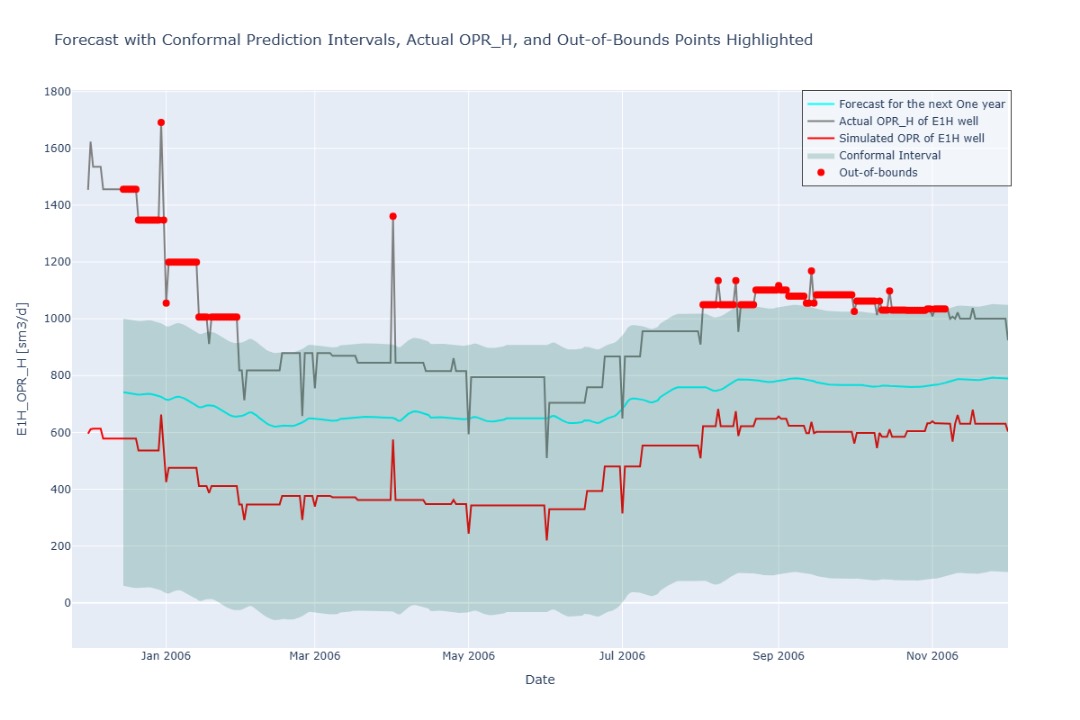}
\caption{LSTM Forecast with CP vs. Actual and Out-of-Bounds Points Well E1H OPR\_H}
\label{fig:figure19}
\end{figure}

\begin{figure}[htb!]
\centering
\includegraphics[width=\linewidth]{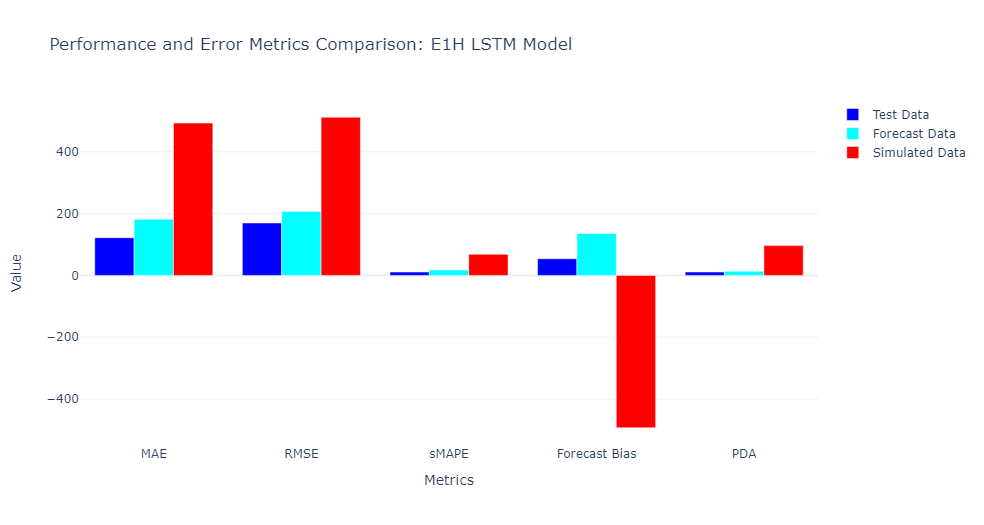}
\caption{Performance and Error Metrics E1H}
\label{fig:figure20}
\end{figure}

The LSTM model, having shown strong performance particularly on PF14, was selected for deployment validation on the Norne E1H well. This cross-validation revealed that the LSTM model is capable of capturing the trend and shape of the actual production reference, matching the trend, magnitude, and shape in the forecast output trial (\cref{fig:figure19}), with a very low percentage of actual OPR outside the 95\% ICP interval (only 7.4\%). This further confirms the effectiveness and potential generalization of the proposed PI-driven ML forecasting approach. The error and performance metrics evaluation for test, forecast, and simulated data versus actual production for E1H are shown in \Cref{tab:table3} and \Cref{fig:figure20}.

\begin{table}[h]
\centering
\caption{Error and Performance Metrics E1H}
\label{tab:table3}
\resizebox{\textwidth}{!}{%
\begin{tabular}{p{0.25\textwidth}p{0.25\textwidth}p{0.25\textwidth}p{0.25\textwidth}}
\hline
\textbf{E1H Metrics} & \textbf{LSTM (Test Data)} & \textbf{LSTM (Forecast Data)} & \textbf{Simulated Data} \\ \hline
\textbf{MAE}           & 121.863 & 182.036 & 492.771  \\ \hline
\textbf{RMSE}          & 169.677 & 206.777 & 511.607  \\ \hline
\textbf{sMAPE}         & 10.974  & 17.260  & 68.294   \\ \hline
\textbf{Forecast Bias} & 53.807  & 135.340 & -492.771 \\ \hline
\textbf{PDA}           & 11.130  & 13.675  & 96.438   \\ \hline
\end{tabular}%
}
\end{table}

\clearpage  

\section{Conclusion}\label{sec4}

Reflecting upon this research, it's obvious that the transition from traditional time series models to machine learning algorithms for hydrocarbon forecasting isn't just a trend but a necessity and rather an imperative requirement. This study demonstrates that Inductive Conformal Prediction (ICP) bridge the gap between machine learning accuracy and actionable uncertainty quantification. By combining ICP with LSTM models, we achieved:  

\begin{itemize}
    \item Robust forecasts: Adapting to nonlinear reservoir dynamics.  
    \item Validated confidence: Ensuring 95\% coverage without distributional assumptions.  
    \item Operational relevance: Enabling risk-aware decisions for reservoir management.  
\end{itemize}

Future work could potentially explore ICP's integration with ensemble models and real-time data streams.

It is paramount that ensuring data quality and integrity before any analysis is key; the process of data preprocessing, imputation (e.g., using KNNImputer), and normalization lays the groundwork for any predictive model. In addition, Hydrocarbon production isn't influenced by time alone. Numerous variables, ranging from inherent geological factors to machinery efficiency, play pivotal roles. Multivariate analysis, hence, emerges as a potentially superior methodology. This study focused on a PI-driven feature set (WPR\_H, BHP\_H, and implicitly GPR\_H through reservoir behavior influence) to predict OPR\_H, demonstrating a robust strategy using reduced dimensionality. The results affirm the importance of considering multiple physics-based influencing factors for more accurate and holistic production forecasting.

Flexibility in model selection and not being tethered to a particular algorithm is crucial. No single model emerges as a one-size-fits-all. The results differ from PF14 ML models to PF12, however model versatility could be achieved by introducing ensembles in future work, and external validations underscored the importance of external validation in asserting model robustness., nevertheless the necessity of generalization always requires model tuning.

In conclusion, while each ML model has its own strengths and weaknesses, their combined insights offer a comprehensive understanding of the oil production rate's time series behavior. The results obtained pave the way for future research in this domain, focusing on refining these models further and incorporating additional variables for even more accurate forecasts, if needed.
\section*{CRediT authorship contribution statement}
\textbf{Mohamed Hassan Abdalla Idris -} Conceptualization, Data Curation, Formal Analysis, Methodology, Software, Writing – original draft
\textbf{Jakub Marek Cebula -} Data Curation, Formal Analysis, Software
\textbf{Shamsul Masum -} Supervision, Validation, Writing – review \& editing
\textbf{Jebraeel Gholinezhad -} Supervision, Validation, Writing – review and editing
\textbf{Hongjie Ma -} Validation

\bibliographystyle{unsrt}
\bibliography{references}  

\end{document}